\def\year{2020}\relax
\documentclass[letterpaper]{article} 
\usepackage{aaai20}  
\usepackage{times}  
\usepackage{helvet} 
\usepackage{courier}  
\usepackage[hyphens]{url}  
\usepackage{graphicx} 
\usepackage{amsmath}
\usepackage{graphicx}  
\title{When Radiology Report Generation Meets Knowledge Graph}

\author{Yixiao Zhang,\textsuperscript{\rm 1} Xiaosong Wang,\textsuperscript{\rm 2} Ziyue Xu,\textsuperscript{\rm 2} Qihang Yu,\textsuperscript{\rm 1} Alan Yuille,\textsuperscript{\rm 1} Daguang Xu\textsuperscript{\rm 2} \\
\textsuperscript{\rm 1} Department of Computer Science, Johns Hopkins University, Baltimore, USA \\
\textsuperscript{\rm 2} NVIDIA Corporation, Bethesda, USA \\
\{wjzzzyx, yucornetto, alan.l.yuille\}@gmail.com, \{xiaosongw, ziyuex, daguangx\}@nvidia.com
}

\begin{document}

\maketitle

\begin{abstract}
\begin{quote}
Automatic radiology report generation has been an attracting research problem towards computer-aided diagnosis to alleviate the workload of doctors in recent years. Deep learning techniques for natural image captioning are successfully adapted to generating radiology reports. However, radiology image reporting is different from the natural image captioning task in two aspects: 1) the accuracy of positive disease keyword mentions is critical in radiology image reporting in comparison to the equivalent importance of every single word in a natural image caption; 2) the evaluation of reporting quality should focus more on matching the disease keywords and their associated attributes instead of counting the occurrence of N-gram. Based on these concerns, we propose to utilize a pre-constructed graph embedding module (modeled with a graph convolutional neural network) on multiple disease findings to assist the generation of reports in this work. The incorporation of knowledge graph allows for dedicated feature learning for each disease finding and the relationship modeling between them. In addition, we proposed a new evaluation metric for radiology image reporting with the assistance of the same composed graph. Experimental results demonstrate the superior performance of the methods integrated with the proposed graph embedding module on a publicly accessible dataset (IU-RR) of chest radiographs compared with previous approaches using both the conventional evaluation metrics commonly adopted for image captioning and our proposed ones.
\end{quote}
\end{abstract}

\section{Introduction}
Interpreting radiology images and writing diagnostic reports is a laborsome and error-prone process for radiologists. Automatic report generation systems can significantly alleviate the burden in the way that candidate reports are provided in natural language for the radiologist to verify. Additionally, learning directly from the free-text reports brings in a huge advantage for adopting data-hungry machine learning paradigms, compared to many other medical image analysis applications that often require large amounts of quality annotations. 

The success of deep learning models on image captioning has motivated a lot of works towards automated radiology report generation for chest x-ray images \cite{yuan2019automatic,li2019knowledge,xue2018multimodal,jing2017automatic,liu2019clinically}. Most of the existing works based on the CNN-RNN Encoder-Decoder framework which has been widely applied in image captioning and visual question answering tasks. Xue et al. \cite{xue2018multimodal} takes multiple image modalities as the input to the encoder. Two-level decoders are included to generate free text paragraphs (multiple sentences) instead of single sentences \cite{jing2017automatic}, while others apply hierarchical generation \cite{krause2017hierarchical} and self-critical sequence training \cite{rennie2017self} to enhance the readability in the radiology reports. 
These models tackled on some aspects of the differences between the natural image captioning and the radiology report generation tasks, e.g., inputs from multiple views, and the fact that a report usually consists of multiple sentences with each one focusing on a specific observation. Nonetheless, one aspect which was not addressed in the previous works is that the correctness of generating clinic-relevant context (positive disease mentions) should be emphasized more than other common words. Furthermore, the medical observations presented in a radiology image are not isolated from each other but may have mutual influence. It is desired that their relationship should be modeled. 

In this work, we build a graph model with prior knowledge on chest findings, which could be injected into the existing models to enhance these two aspects. In this graph, disease findings are defined as nodes and related findings are closely connected so that they can influence each other during the graph propagation and aggregation. We incorporate this graph into the deep neural network to learn dedicated features for each node on the graph. These graph features are later used for the classification and report generation. Specifically, the graph embedding module is computed after a CNN feature extractor, and an attention mechanism is designed to compute initial node features from CNN features. Then, graph convolutions are conducted to propagate features over the chest abnormality graph. As the output, a linear classifier for classification and a multi-level decoder module for report generation are connected to the graph convolution layers respectively. We decompose the learning process into two stages. First, we train a multi-label classification network where each class corresponds to an observation, therefore, also corresponds to a node on the composed graph. The model is encouraged to learn discriminatory features for classifying disease findings. After training the classification network, a decoder that consists of a topic level LSTM and a word level LSTM is trained to generate reports. The decoder learns to attend to different findings on the graph, and focuses on one concept at each sentence.

\begin{figure*}
    \centering
    \includegraphics[width=0.75\linewidth]{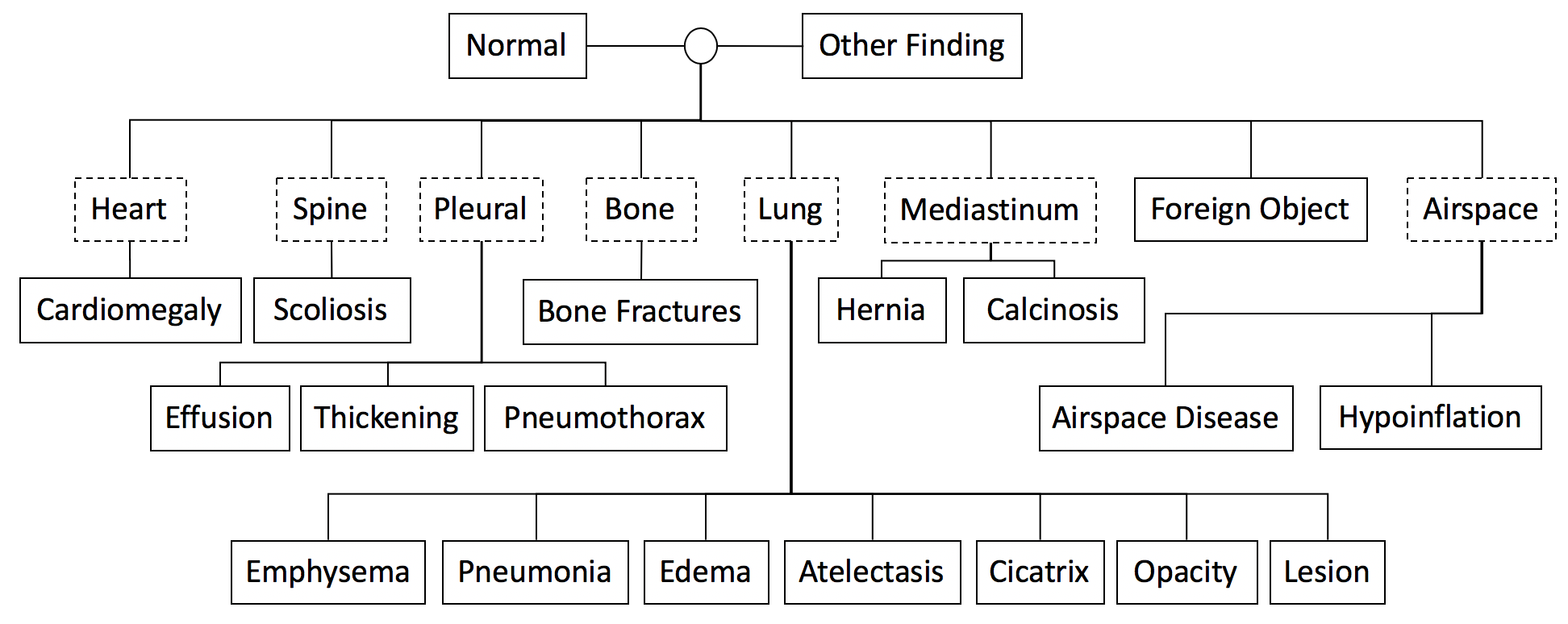}
    \caption{An illustration of all the findings and their grouping in our composed graph. The solid boxes are classes which have corresponding nodes in graph. The dotted boxes are organs or tissues and are not part of target classes. Classes linked to the same organ or tissue are connected to each other in the graph. The root here represents the global node in the graph.}
    \label{fig:graph}
\end{figure*}

In addition, the Bilingual Evaluation Understudy Scores (BLEU-N) \cite{papineni2002bleu} together with many other evaluation metrics, e.g., ROUGE \cite{lin2004rouge}, CIDEr \cite{vedantam2015cider} have been widely used for measuring the quality of generated image captions via matching the occurrence of N-gram against the ground truth. In the matching, every individual word contribute equally to the final evaluation score. Nevertheless, they may not demonstrate the true accuracy when they are used for measuring the quality of medical image reporting, since radiologists often report to exclude many diseases (either commonly diagnosed or intended by the physicians) using negation expressions, e.g., no, free of, without, etc. For example, "there are increased interstitial markings without evidence of focal airspace disease" and "there are increased interstitial markings with evidence of focal airspace disease" are two sample reports. They have a high BLEU-1 score of 0.9 but their meanings are actually opposite. Furthermore, the correct detection rate of disease mentions may be overwhelmed by the accuracy of other non-significant words, e.g. stop words. Based on these observations, we believe that a new evaluation metric which focuses on the correctness of detected diseases in the report should be designed. Here, we propose a new evaluation metric, named the Medical Image Report Quality Index (MIRQI), to accent the correctness of both positive and negative disease mentions and their associated attributes in the generated reports.

We evaluate our work using the publicly accessible IU-RR dataset \cite{demner2015preparing}. The performance of our model in both classification and report generation tasks is compared with previous arts in both quantitative and qualitative manner. In classification, our model performs better in most of the categories and achieves 2\% Area Under Curve (AUC) improvement on average. In report generation, our model obtains better or equivalent performance in conventional evaluation metrics, and at the meantime scores significantly higher in the proposed MIRQI metrics. It indicates that utilizing graphs with prior knowledge is helpful to generate more accurate reports from both the language and clinical correctness perspectives.

\section{Related Works}

In diagnostic radiology, radiologists read radiology images of patients, identify abnormalities or diseases, and record their findings or conclusions in reports. A report typically consists of many sections, e.g., comparison, indication, findings and impression. Findings are detailed descriptions of all kinds of observations in the image, including both normal and abnormal ones. Impression, on the other side, is a summary of observations, which usually only has one or two sentences. Similar to previous works, we are aimed to generate findings and impression parts together.

As previously mentioned, many works have explored deep learning based methods for report generation. Wang et al. \cite{wang2018tienet} proposed a text-image embedding network to jointly learn the textual and image information for both the classification and image reporting task.  Towards a similar direction, Jing et al. \cite{jing2017automatic} presented a multi-task framework which first learns to predicts medical tags then generate text description, in which they employed a co-attention mechanism over both visual features and textural embedding. Besides, hierarchical multiple-level Long short-term memory (LSTM) units are integrated as the decoder. Xue et al. \cite{xue2018multimodal} proposed a recurrent generation model, where the generation of a sentence is based on both the visual features and the encoded feature of the previous sentence. They also fused visual information in multiple views by concatenating their CNN features. Liu et al. \cite{liu2019clinically} applied self-critical sequence training \cite{rennie2017self} based on reinforcement learning to optimize a clinically coherent reward, which focuses on the correct mention of disease keywords. Yuan et al.\cite{yuan2019automatic} explored many ways of fusing frontal and lateral view features, and used attention over medical concepts which are extracted from Medical Text Indexer.

In our proposed framework, we followed some successful practices of the previous works, including the fusion of features from frontal and lateral views, and a two-level decoder for topic and sentence generation individually. Our main goal is to demonstrate the performance gain from the incorporation of the graph module with prior knowledge which allows the interaction of representative features between findings.


As many existing works \cite{yao2018exploring,liang2018symbolic,chen2018iterative,norcliffe2018learning,hu2019language}, we used the graph convolution as a means of message passing and node interaction. However, the way that we applied the graph modeling differs from others. First, radiology images exhibits less variability than natural images in terms of overall contents. We use a universal graph for all images, while for natural images scene graphs are constructed based on object detection and relationship prediction, and can vary from image to image. Second, there is no available ground truth bounding boxes to locate findings in radiology images, which requires new ways to notate findings and initialize dedicated node features using graphs.

\section{Graph Construction with Prior Knowledge}
Graph structures are often used to represent entities and their relationships. In our work, we compose a graph that covers the most common abnormalities or findings in chest X-rays. 
Each node in the graph represents one of the findings and is denoted by disease keywords. Apart from 'normal', 'other' and 'foreign object', all other findings are grouped by the organ or body part that they relate to. Figure~\ref{fig:graph} illustrates the disease keywords and their grouping in our setting. Dotted boxes indicate the group categories as virtual nodes. For findings grouped together, we connect their nodes with bidirectional edges. Additionally, we use a separate node to represent the global information, and connect this node to all other nodes.

We designed this graph based on prior knowledge from clinical studies \cite{chest_tutorial}. For example, abnormalities on the same body part may have strong correlation with each other and share many features, while relations between abnormalities of different organs should be minor. However, we note that more sophisticated relationships could be annotated with more complex graph structures, and our model is not limited to the underlying graph. Disease categories utilized in previous works, e.g., ChestX-ray8 \cite{wang2017chestx} and CheXpert \cite{irvin2019chexpert} are also considered here. Finally, we obtained 20 keywords (categories) in the defined chest abnormality graph, which will be utilized to facilitate our classification and report generation applications in the following sections.

\begin{figure*}[!t]
    \centering
    \includegraphics[width=0.8\linewidth]{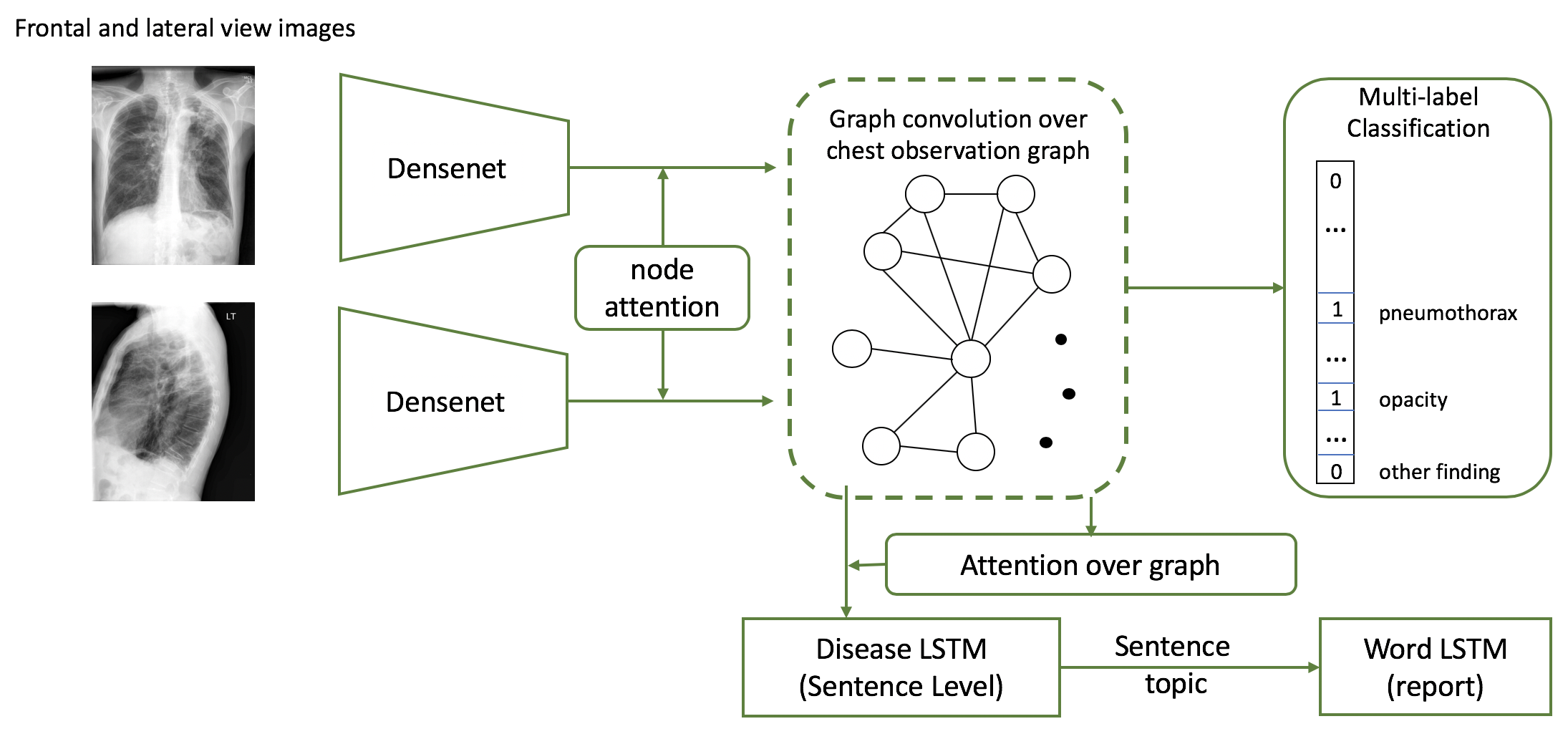}
    \caption{Overview of the proposed framework. Graph node features are extracted from CNN features, followed by graph convolution layers. There are two branches after graph convolution: one for classification and one for report generation.}
    \label{fig:flowchart}
\end{figure*}

\section{Multi-Label Classification via Graph Embedding}
As shown in Figure~\ref{fig:flowchart}, DenseNet-121 \cite{huang2017densely} pre-trained on CheXpert \cite{irvin2019chexpert} was adopted as the backbone of our proposed network. For both tasks, images of frontal and lateral views are inputted to the backbone CNN model, then their features are fed to the graph embedding module through an attention mechanism. After that, the graph features of both views are concatenated. The framework then branches into a multi-label classifier and a report generation decoder. The classification branch was trained first and remain fixed during the training of the report generation decoder.

The targets of classification branch are defined as the finding categories in our graph. Each node in the graph corresponds to a finding category except the global node. During the training and testing of this classifier, the number of nodes in graph are fixed. We initialize all the node features using an attention mechanism on CNN features. Then, graph convolution layers are applied to propagate messages over the graph. Finally, the node features are used to produce class predictions, which are elaborated in details as follows.

\subsection{Node Feature Initialization}
After the block 4 in DenseNet-121, we employ a spatial attention module (node attention module in Figure~\ref{fig:flowchart}) upon the output activation. The attention map computation is implemented using a Convolution layer with filter size of $1\times1$ followed by a softmax layer over the spatial locations, where the number of channels equals to the number of finding classes. Then, the initial feature of a node in the graph is obtained as the attention-weighted sum of the activation, where attention weights come from the corresponding channel. 
The feature of the global node is initialized with the output of global average pooling. In this way, each node on the graph learns to attend to a different spatial area, and would learn its own dedicated feature for the corresponding finding.

\subsection{Graph Convolution}
After obtaining the initial node features, the graph convolution is used to propagate information on the graph. We mainly followed the graph convolution operation in \cite{kipf2016semi} with some modifications. In general, the graph convolution can be expressed as
\begin{equation}
F^{l+1} = update(F^l, message(F^l, A))
\end{equation}
where $F^l$ is the node features in the $l$-th layer, $F^{l+1}$ is the node features in the ($l+1$)-th layer, $message$ is a function to generate and aggregate messages based on the features $F^l$ and the normalized adjacency matrix $A$, and $update$ is a function to update node features based on messages. In this work, we implemented the graph convolution as
\begin{align}
    m & = ReLU(BN(Conv1d(F^l)A)) \\
    F^{l+1} & = ReLU(BN(Conv1d(concat(F^l, m))))
\end{align}
where $A$ is the normalized Laplacian of the adjacency matrix, $m$ is the aggregated message for each node. In each graph convolution layer, messages are computed using 1d convolution for both incoming and outgoing edges. Then, messages from neighbors are aggregated by multiplying the normalized Laplacian matrix. Finally, current node features as well as messages are used to update the node features through another 1d convolution layer. Batch Normalization ($BN$) and $ReLU$ layers are added after each convolution layer and residual connections are also introduced between layers.

\subsection{Loss Functions}
At the end of graph convolution layers, global average pooling was applied to obtain a graph level feature, then a fully-connected layer with $Sigmoid$ activation was used to predict probabilities for each finding as a multi-label classification task. We used weighted binary cross entropy loss for the training considering the positive/negative imbalance in the dataset. However, using this loss only is sufficient to regularize what features each node should learn and which part of the feature map it should attend to. Therefore, we added an auxiliary loss to the node attention module. For each node, after obtaining its initial features from the attention module, we added a fully-connected layer with $sigmoid$ activation which served as an auxiliary classifier. Each node would be enforced to represent a specific finding and determine the existence of it. In such way, the nodes are distinguishable from each other, and are guided to attend to different areas of the image for different disease categories.

\section{Report Generation via Graph Embedding}
After training the multi-label classification model, we fixed the parameters in both the CNN backbone and the graph embedding module, and appended after the graph embedding module with a two-level decoder to generate reports. Our decoder is composed of two level of recurrent units, one at topic level and another at word level. The choice of a two-level decoder is inspired by the observation that medical reports usually constitutes multiple sentences with each focusing on one topic. The recurrent units could vary according to different applications, e.g., LSTM and Gated Recurrent Unit (GRU). We experiment with LSTM for our applications.

\subsection{Topic Generation}
\subsubsection{Attention over Graph Embedding}
The input to the topic-LSTM is a context vector computed from the graph embedded features. We utilize another attention mechanism here to obtain the context vector as a weighted summary of the graph node features for different topics. Given the hidden state of the topic-LSTM $h_{s,t-1}$ from time $t-1$ and the graph embedded features $E=\{e_i, i=1,...,N\}$, the attention weight for each node is computed using a two-layer network with $softmax$ activation.
\begin{align}
    a_i & = W_a \tanh{(W_v e_i + W_s h_{s,t-1})} \\
    \alpha_i & = softmax(a_i)
\end{align}
where $W_a, W_v, W_s$ are parameters, and \{$\alpha_i$\} are attention weights on each node $i$. The context vector is then computed as
\begin{equation}
    v_t = \sum_i \alpha_i e_i
\end{equation}
Therefore, the attention module takes information about what have been predicted (the last hidden state) and gives what should be focused on for the next sentence (the context vector). Since the attention is applied over the graph nodes rather than the CNN features, the generated topic would focus more on the finding concepts that it is attending to. At beginning, the hidden state of the topic-LSTM is initialized by the global averaged CNN features. Then, its hidden state is updated for each sentence and remains steady during the prediction of one sentence. 


\subsection{Sentence Generation}
The topic-LSTM outputs topic vectors $s_t$, which are feed into the word-LSTM. The word-LSTM also takes the context vector $v_t$ from the graph attention module, and predicts the detailed sentence in a word by word fashion. Note that both the topic vector and the context vector are used in updating word-LSTM gates and states according to the following functions
\begin{align}
    i_{w,\tau} & = \sigma(W_{si} s_t + W_{vi} v_t + W_{hi} h_{w,\tau-1}) \\
    f_{w,\tau} & = \sigma(W_{sf} s_t + W_{vf} v_t + W_{hf} h_{w,\tau-1}) \\
    g_{w,\tau} & = \tanh(W_{sg} s_t + W_{vg} v_t + W_{hg} h_{w,\tau-1}) \\
    o_{w,\tau} & = \sigma(W_{so} s_t + W_{vo} v_t + W_{ho} h_{w,\tau-1}) \\
    c_{w,\tau} & = f_{w,\tau} * c_{w,\tau-1} + i_{w,\tau} * g_{w,\tau} \\
    h_{w,\tau} & = o_{w,\tau} * \tanh(c_{w,\tau})
\end{align}
where the subscript $w$ stands for `word' and $\tau$ stands for time step. $i,f,o$ are the input gate, forget gate, output gate respectively. $c$ is cell state and $h$ is hidden state. All the $W_*$ are parameters.

\section{Quality Evaluation by Graph Matching}
\begin{figure}
	\centering
	\includegraphics[width=\linewidth]{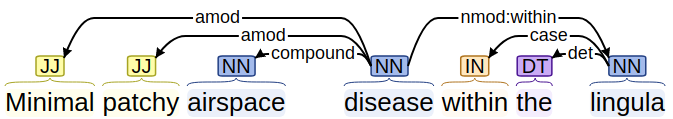}
	\caption{It illustrates a dependency parsing of a sample sentence from the report. "minimal", "patchy", and "lingula" are extracted as the attributes of "airspace disease".}
	\label{fig:dep_graph}
\end{figure}

 In the proposed MIRQI evaluation, both of the paired reports (the ground truth report and generated one) will be processed with disease word extraction, negation/uncertainty extraction, and attributes extraction based on dependency graph parsing. We adopted a similar method proposed in NegBio \cite{peng2018negbio,wang2017chestx} and CheXpert \cite{irvin2019chexpert} labeling toolkit for entity extraction and rule-based negation detection. It also considers synonyms and variations of disease words during the searching and represents the findings with representative disease words (as listed in the defined abnormality graph). The extracted disease keywords will compose a graph for each individual report, which is indeed a sub-graph of the constructed chest abnormality graph stated before. Additionally, we process each sentence in the report with the Stanford parser \cite{chen2014fast} to generate the dependency graph (as an example illustrated in Figure~\ref{fig:dep_graph}). A list of disease keywords' child nodes could then be extracted as the attributes, including adjectival modifier (amod), nominal modifier (vmod), negative (neg), direct object (dobj), nominal subject (nsubj), and compound. These attributes represent the features of disease, such as severity, size, shape, body parts, and many other aspects. 
	
Given a pair of sub-graphs, one from prediction and the other from ground truth, we compute the recall (MIRQI-r) of disease mentions and associated attributes in a node-by-node fashion as,
\begin{equation}
\mbox{MIRQI-r} = w_{pos}*\dfrac{TP}{TP+FN}+w_{neg}*\dfrac{TN}{TN+FP}
\end{equation}
where True Negative ($TN$), False Positive ($FP$) and False Negative ($FN$) are computed by matching paired graphs in a node by node fashion. True Positive ($TP$) will additionally include the correct hits of attributes for each positive disease mentions,
\begin{equation}
TP=(1-w_{attr})*TP_{keywords}+w_{attr}*TP_{attributes}
\end{equation}
where $w_{pos}$ and $w_{attr}$ weight the contribution of positive mentions and attributes, and $w_{neg}=1-w_{pos}$. In a similar fashion, we define the precision (MIRQI-p) as,
\begin{equation}
\mbox{MIRQI-p} = w_{pos}*\dfrac{TP}{TP+FP}+w_{neg}*\dfrac{TN}{TN+FN}
\end{equation}
and the $F_{1}$-measure (MIRQI-F1) score as,
\begin{equation} 
\mbox{MIRQI-F1}=\frac{\mbox{MIRQI-r}*\mbox{MIRQI-p}}{\mbox{MIRQI-r}+\mbox{MIRQI-p}}
\end{equation}

\section{Experiments and Results}

In this section, we report several experiments that explored and validated the advantage of including graph embedding module in radiology abnormality classification and report generation. First, we reveal more details of constructing the prior knowledge graph about chest findings/abnormalities in our experiments. Second, we evaluate the performance of incorporating graph embedding module into a strong baseline DenseNet-121 for multi-label abnormality classification. Finally, we evaluate the report generation decoder based on the learned graph embedded features, which shows better performance under both the conventional metrics as well as the proposed MIRQI scores.

\subsubsection{Experimental Setting}
We used the publicly accessible dataset IU-RR \cite{demner2015preparing} for evaluating all our models. The dataset contains 3955 radiology reports, each associated with one frontal view chest x-ray image and optionally one lateral view image. A report mainly consists of comparison, indication, findings and impression sections, where findings is a list of findings and impression is the overall diagnosis. For our experiments, we only include cases with both frontal and lateral views, and with complete findings and impression sections in the report. This results in totally 2902 cases and 5804 images.

Input image size is $512\times512$, and the feature map from DenseNet-121 block 4 is $1024 \times 16 \times 16$. We randomly crop a $512 \times 512$ region with padding if needed, and no other data augmentation is used for all experiments.

\begin{table*}[t]
    \small
    \centering
    \begin{tabular}{c|c|c|c|c|c|c|c}
         & average & normal & cardiomegaly & scoliosis & FB & effusion & thickening \\
        \hline
        ChestXray8\cite{wang2017chestx} & 0.719 & - & 0.803 & - & - & 0.890 & - \\
        TieNet\cite{wang2018tienet} & 0.779 & 0.747 & 0.847 & - & - & 0.899 & - \\
        Densenet\cite{irvin2019chexpert} & 0.778 & 0.795 & 0.866 & \textbf{0.664} & \textbf{0.695} & 0.921 & \textbf{0.733} \\
        Densenet+KG & \textbf{0.792} & \textbf{0.807} & \textbf{0.913} & 0.663 & 0.671 & \textbf{0.942} & 0.728 \\
        \hline
         & pneumothorax & hernia & calcinosis & emphysema & pneumonia & edema & atelectasis \\
        \hline
        ChestXray8\cite{wang2017chestx} & 0.631 & - & - & 0.675 & 0.642 & 0.799 & - \\
        TieNet\cite{wang2018tienet} & 0.709 & - & - & 0.792 & 0.731 & 0.879 & - \\
        Densenet\cite{irvin2019chexpert} & 0.824 & 0.860 & \textbf{0.676} & \textbf{0.892} & 0.844 & 0.897 & 0.788 \\
        Densenet+KG & \textbf{0.843} & \textbf{0.884} & 0.669 & 0.890 & \textbf{0.863} & \textbf{0.931} & \textbf{0.833} \\
        \hline
         & cicatrix & opacity & lesion & AD & hypoinflation & MD & other \\
        \hline
        ChestXray8\cite{wang2017chestx} & - & - & 0.647 & - & - & - & - \\
        TieNet\cite{wang2018tienet} & - & - & \textbf{0.658} & - & - & - & - \\
        Densenet\cite{irvin2019chexpert} & \textbf{0.742} & 0.796 & 0.597 & 0.830 & 0.768 & 0.775 & 0.595 \\
        Densenet+KG & 0.734 & \textbf{0.803} & 0.643 & \textbf{0.857} & \textbf{0.775} & \textbf{0.805} & \textbf{0.596}
    \end{tabular}
    \caption{Comparison of multi-label classification models. AUC scores are computed for the overall average and on each individual category. FB: fractures bone. AD: airspace disease. MD: medical device}
    \label{tab:class}
\end{table*}

\begin{table*}[t]
    \centering
    \begin{tabular}{c|c|c|c|c|c|c|c|c|c}
         & BLEU-1 & B-2 & B-3 & B-4 & CIDEr & ROUGE & MIRQI-r & MIRQI-p & MIRQI-F1 \\
        \hline
        CoAtt\cite{jing2017automatic} &   0.455 & 0.288 & 0.205 & 0.154 & 0.277 & 0.369 & - & - &- \\
        KER\cite{li2019knowledge} &   0.455 & 0.304 & 0.210 & - & 0.318 & 0.335 & - & - &- \\
        \hline
        TieNet\cite{wang2018tienet} &   0.330 & 0.194 & 0.124 & 0.081 & - & 0.311 & - & - &- \\
        CARG\cite{liu2019clinically} &   0.359 & 0.237 & 0.164 & 0.113 & - & 0.354 & - & - &- \\
        \hline
        \hline
        SAT\cite{xu2015show} & 0.433 & 0.281 & 0.194 & 0.138 & 0.320 & 0.361 & 0.478 & 0.479 & 0.471 \\
        SentSAT\cite{yuan2019automatic} & \textbf{0.445} & 0.289 & 0.200 & 0.143 & 0.268 & 0.359 & 0.470 & 0.472 & 0.462 \\
        SentSAT+KG & 0.441 & \textbf{0.291} & \textbf{0.203} & \textbf{0.147} & \textbf{0.304} & \textbf{0.367} & \textbf{0.483} & \textbf{0.490} & \textbf{0.478}
    \end{tabular}
    \caption{Comparison of report generation models on both image captioning metrics and the proposed MIRQI metrics. Note: the results in the top 2 sections are reported in \cite{li2019knowledge} and \cite{liu2019clinically} separately with different experimental settings.}
    \label{tab:report}
\end{table*}

We included 20 finding keywords as disease categories, which is more complete than the previous works. These keywords cover the most common findings of organs or areas in the chest. To get ground truth labels for classification, we detect the keywords in the Mesh part of the reports which lists findings in a formatted way. 

To evaluate our models, we employed stratified five-fold cross validation which ensures that the number of samples in each fold is roughly the same for every finding category. The split of data in the same category are totally random. The average score on five folds are reported.

We tokenize all the words in the reports and drop infrequent tokens with frequency less than three. This results in 1524 unique tokens, including four special tokens $<$pad$>$, $<$start$>$, $<$end$>$ and $<$unknown$>$. The findings and impression sections are concatenated as the ground truth report.

\subsubsection{Evaluation Metrics}
For the quantitative evaluation, we employed the AUC of Receiver Operating Characteristic (ROC) curve to measure the classification performance. We used some common metrics for image captioning including BLEU, ROUGE, CIDEr scores as well as the proposed MIRQI metrics to evaluate the reports. $w_{pos}$ is set to 0.8 and $w_{attr}$ is set to 0.2.

\subsubsection{Prior Knowledge Graph Construction}
As mentioned above, we extracted 20 class keywords from the reports, which corresponds to the nodes in the chest abnormality graph. Abnormalities on the same organ may correlate with each other. Therefore, we divided the classes into groups by the organs to which they are related, and connected the nodes whose corresponding classes are in the same group. We added a node which connects to all the other nodes, thus associating all groups of nodes. In our design, this node captures the global visual information of the radiology images.

\subsection{Results on Multi-label Classification}
For classification, we use Densenet \cite{irvin2019chexpert} as our baseline. It is pretrained on the CheXpert dataset. We replaced the last fully-connected layer with a multi-label classification layer and finetune the whole model on the IU-RR dataset. Our proposed model is notated as Densenet+KG, where the attention and graph convolution layers are appended to the Densenet backbone. The AUC scores on average and for each class are listed in Tabel~\ref{tab:class}. We also included Several previous works for comparison, i.e. ChestXray8 \cite{wang2017chestx} and TieNet \cite{wang2018tienet}.

For most of the classes, our proposed model achieves higher or equivalent AUC scores. On average, the improvement is about 2\%. Since the overall settings are identical for the baseline and our proposed model, the improvement solely comes from the use of the chest abnormality graph. A possible explanation is that the model learns disentangled concepts for each node on the graph, and message passing through graph convolution allows the interaction between the prediction of correlated classes.

\begin{figure*}[t]
    \centering
    \includegraphics[width=\linewidth]{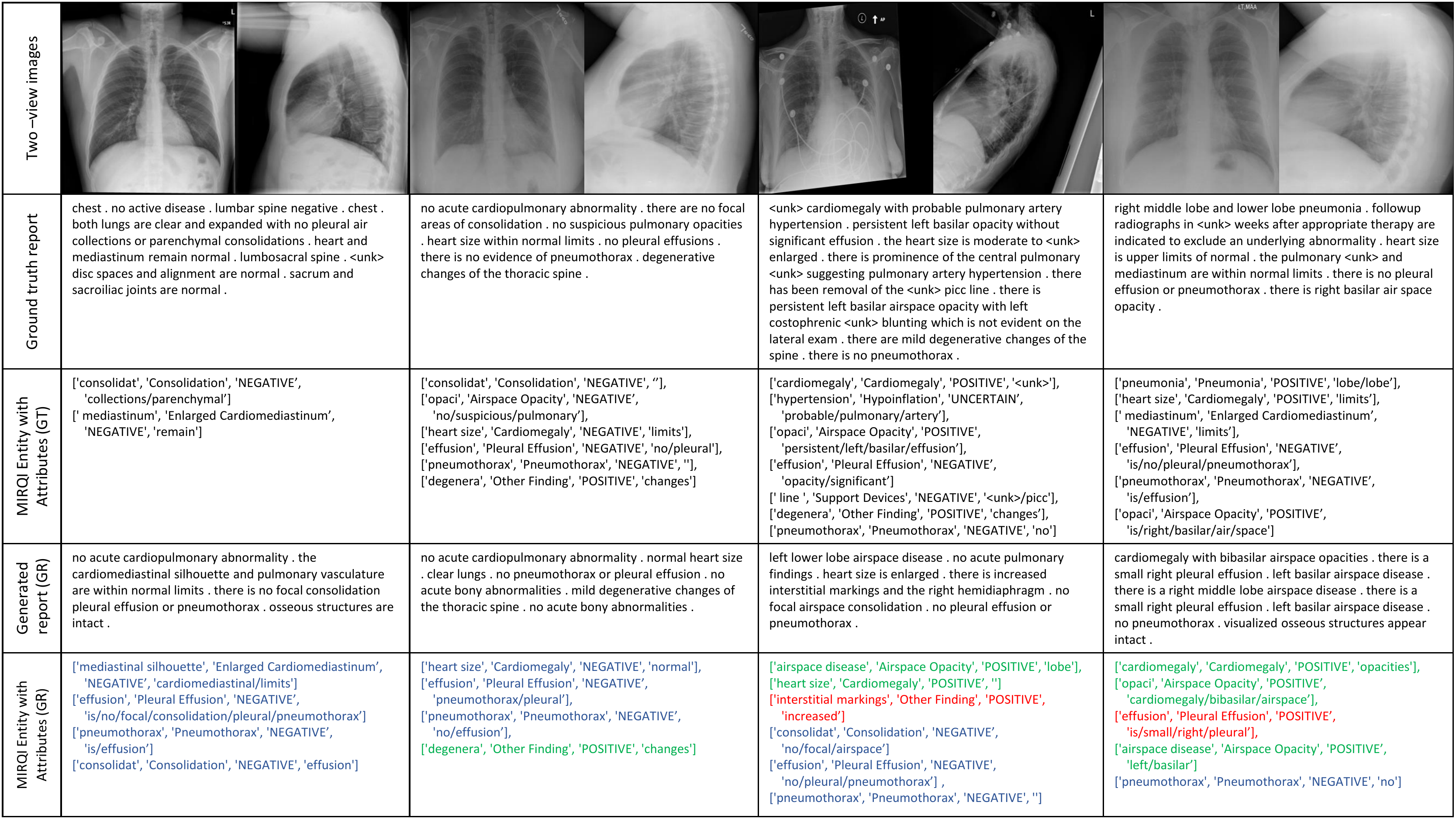}
    \caption{4 sample cases with two-view images on the top, the ground truth report and generated ones on the 2nd and 4th rows. The 3rd and 5th rows illustrated the extracted disease keywords and attributes from GT and GR individually. Text in Blue: true negative; Green: true positive; Red: false positive. Each MIRQI entity contains [`word', `category', 'negation', 'attributes'].}
    \label{fig:vis}
\end{figure*}

\subsection{Results on Report Generation}
We compare our model with several previous works on radiology report generation. The first is the classic Show, Attend and Tell work (SAT) \cite{xu2015show}. It has only one level of recurrent units in the decoder. We further extend the SAT model with additional sentence-level LSTM (SentSAT) (similar to the multi-level LSTM framework in \cite{yuan2019automatic} but without medical concept injection). The difference between SentSAT and our model is that the former uses attention over CNN features to obtain the context vector, while the latter first extracts chest abnormality graph features from the CNN features, propagates information on the graph, and then obtain the context vector using attention over graph node features. All other parts of the models are the same, which makes it a fair comparison. We represent our proposed model as SentSAT+KG. We also include previous works that reported results on dataset IU-RR, while please note that these evaluations may result from different experiment settings, data splits, and preprocessing on the corpus, which we find have large impact on the performance. 

Table~\ref{tab:report} shows the performance of all three models on both image captioning metrics and the proposed MIRQI-r (Recall), MIRQI-p (Precision) and MIRQI-F1 metrics. Our proposed model performs better than SAT and SenSAT in most of the language metrics. This suggests that attention over the chest abnormality graph is an alternative of attention over CNN feature maps for text generation tasks, as long as the graph covers the needed concepts. Besides, our model achieves 1.3\%-1.8\% improvement on the MIRQI metrics, which indicates that the generated reports are more accurate in detecting diseases. Our proposed method also achieves equivalent or higher scores compared to CoAtt, KER, TieNet and CARG on the same IU-RR dataset although it may not be a fair comparison due to different experimental settings. Only BLEU-N scores, CIDEr and ROUGE are reported in this case. All these metrics reflects some aspects of methods' performance, e.g., BLEU is more close to precision and CIDEr leans to recall, while the proposed MIRQI metric are designed to cover both sides and focus more on the clinical relevant texts.

\subsection{Qualitative Results}
In Figure~\ref{fig:vis}, we visualized four sets of sample images along with their ground truth and generated reports. The extracted disease findings and their attributes from MIRQI are also listed. The one on the left illustrates a normal case. The model is able to generates negative mentions correctly and also add in two more negative mentions, which happens often in all 4 cases and will not hurt the overall correctness of generated reports. In the rest 3 cases, our proposed method demonstrates its capability of generating both correct positive and negative mentions. For example, `Airspace opacity' and `Cardiomegaly' are accurately reported in the third case, while the model also generates a false mention of `other finding'. Furthermore, one interesting point about our proposed model is that it intends to output similar sentences for the same disease findings for multiple times. For example, the `airspace disease' are repeated in the far-right case. In such cases, we believe the topic attention mechanism has play an role in emphasizing more confident findings topics (from the classification point of view). 

\section{Conclusions}
In this paper, we propose to use the chest abnormality graph with prior knowledge of chest X-ray to assist radiology report generation. 
Attention mechanism and graph convolution are adapted to learn the graph embedded features. Then, we are capable of utilizing the disentangled features of the graph nodes to boost classification and report generation. Additionally, we proposed MIRQI metrics to examine the correctness of positive and negative disease mentions in the report. Our model outperforms the previous approaches both in language metrics and the MIRQI metrics. Our model is not limited to the specific structure of the pre-contructed graph ,and more sophisticated graph structures (with more detailed disease relationship modelling) can be considered in the future. Importantly, we will make our code (both the model and metrics) and data split public available to promote a fair comparison for the future evaluation.


\bibliography{papers}

\end{document}